\documentclass{article}

 \usepackage[final]{neurips_2025_ml4ps}

\usepackage[utf8]{inputenc} 
\usepackage[T1]{fontenc}    
\usepackage{hyperref}       
\usepackage{url}            
\usepackage{booktabs}       
\usepackage{amsfonts}       
\usepackage{nicefrac}       
\usepackage{microtype}      
\usepackage{xcolor}         

\usepackage{caption}
\usepackage{subcaption}
\usepackage{amsmath,amssymb,amsfonts}
\usepackage{algorithm}
\usepackage{algorithmic}
\usepackage{graphicx}
\usepackage{textcomp}

\usepackage{microtype}
\usepackage{enumitem}
\setlist[itemize]{noitemsep, topsep=0pt, parsep=0pt, partopsep=0pt}
\usepackage{bm}
\usepackage{makecell} 
\usepackage{bbold}

\usepackage{mathtools}
\mathtoolsset{showonlyrefs}
\usepackage{amsthm}

\theoremstyle{plain}

\theoremstyle{definition}

\theoremstyle{remark}

\title{Learning Solution Operators for Partial Differential Equations via Monte Carlo-Type Approximation}

\author{%
  Salah Eddine Choutri           \\
  NYUAD Research Institute       \\
  New York University Abu Dhabi  \\
  United Arab Emirates, UAE      \\
  \texttt{sc8101@nyu.edu}        \\
  \And \hspace{3cm}
  Prajwal Chauhan                \\ \hspace{3cm}
  Engineering Division          \\ \hspace{3cm}
  New York University Abu Dhabi  \\ \hspace{3cm}
  United Arab Emirates, UAE      \\ \hspace{3cm}
  \texttt{pc3377@nyu.edu}        \\ \hspace{3cm}
  \And \hspace{-2cm}
  Othmane Mazhar                  \\ \hspace{-1cm}
  Laboratoire de Probabilit\'es, Statistique et Mod\' elisation \\ \hspace{-1cm}
  Sorbonne University \& Universit\'e Paris Cit\'e      \\ \hspace{-2cm}
  Paris, France                       \\ \hspace{-2cm}
  \texttt{omazhar@lpsm.paris}         \\ \hspace{-2cm}
  \And \hspace{0.6cm}
  Saif Eddin Jabari              \\ \hspace{0.6cm}
  Engineering Division         \\ \hspace{0.6cm}
  New York University Abu Dhabi  \\ \hspace{0.6cm}
  United Arab Emirates, UAE      \\ \hspace{0.6cm}
  \texttt{sej7@nyu.edu}          \\ \hspace{0.6cm}
   }

\begin{document}

\maketitle

\begin{abstract}
The Monte Carlo-type Neural Operator (MCNO) introduces a lightweight architecture for learning solution operators for parametric PDEs by directly approximating the kernel integral using a Monte Carlo approach. Unlike Fourier Neural Operators, MCNO makes no spectral or translation-invariance assumptions. The kernel is represented as a learnable tensor over a fixed set of randomly sampled points.
This design enables generalization across multiple grid resolutions without relying on fixed global basis functions or repeated sampling during training.
Experiments on standard 1D PDE benchmarks show that MCNO achieves competitive accuracy with low computational cost, providing a simple and practical alternative to spectral and graph-based neural operators.
\end{abstract}

\section{Introduction}
Neural operators extend the success of neural networks from finite-dimensional vector mappings to infinite-dimensional function spaces, enabling the approximation of solution operators for partial differential equations (PDEs). These models learn mappings from problem inputs, such as coefficients or boundary conditions, to PDE solutions, allowing rapid inference and generalization across parameterized problem families. Early architectures like DeepONet~\cite{lu2021learning} use branch and trunk networks, while Fourier Neural Operators (FNO)~\cite{li2021fourier} leverage spectral convolutions and FFTs for efficient operator approximation on uniform grids. Wavelet-based models such as MWT~\cite{gupta2021multiwavelet} and WNO~\cite{tripura2022wavelet} improve spatial localization, and graph-based approaches like GNO~\cite{li2020neuralb} extend operator learning to irregular domains.

We propose the Monte Carlo-type Neural Operator (MCNO), a lightweight architecture that approximates the integral kernel of a PDE solution operator via Monte Carlo sampling, requiring only a single random sample at the beginning of training. By avoiding spectral transforms and deep hierarchical architectures, MCNO offers a simple yet efficient alternative that balances accuracy and computational cost. We evaluate MCNO on standard 1D PDEs, including Burgers’ and Korteweg–de Vries equations, using benchmark datasets, and show competitive performance against existing neural operators while maintaining architectural simplicity. Our contributions include the MCNO design, feature mixing for spatial and cross-channel dependencies, an interpolation mechanism for structured grids and benchmark evaluation.

\section{Monte Carlo-type Neural Operator (MCNO)}
Neural operators provide a framework for learning mappings between infinite-dimensional function spaces, enabling efficient approximation of solution operators for parametric PDEs~\cite{kovachki2023neural}. For a generic PDE, the true solution operator $G^\dagger$  is a mapping that takes an input $a$ to the corresponding PDE solution $u$ s.t $G^\dagger: \; a \mapsto u = G^\dagger(a)$. For many inputs, evaluating $G^\dagger$ with classical solvers can be prohibitively expensive. Given a dataset $\{(a_j, u_j)\}_{j=1}^N$ with $u_j = G^\dagger(a_j)$, the goal is to learn a parameterized approximation $G_\theta \approx G^\dagger$ by minimizing a suitable loss. A neural operator adopts iterative architectures to compute such approximate operator $G_\theta$, to this end it lifts the input to a higher-dimensional representation $v_0(x) = P\big(a(x)\big), \ x \in \mathcal{D} \subset \mathbb{R}^d$ via a local map $P$. It then, updates it through a linear transformation via matrix $W$ and a kernel-integral transformation $(\mathcal{K}_\phi v)(x)$ with nonlinear activations $\sigma$. The $(t+1)\text{th}, \ t \in \mathbb{N},$ neural operator layer is expressed as:
\[
v_{t+1}(x) = \sigma \left( Wv_t(x) + (\mathcal{K}_\phi v_t)(x) \right), \quad x \in \mathcal D,
\]
with
\[
(\mathcal{K}_\phi v_t)(x) = \int_D \kappa_\phi(x, y, a(x), a(y)) v_t(y) dy,
\]
where \(\kappa_\phi\) is a learned kernel inspired by the Green function commonly involved in the solution of linear PDEs. A final projection $Q$ maps the final features back to the target function space
\begin{equation}
  \big(G_\theta(a)\big)(x) = Q\big(v_T(x)\big), \qquad x \in \mathcal{D}.
\end{equation}
MCNO follows this framework but replaces the integral with a Monte Carlo estimate over a fixed set of sampled points, i.e.,
\begin{equation}
(\hat{\mathcal{K}}_{N,\phi} v_t)(x) = \frac{1}{N} \sum_{i=1}^N \kappa_\phi(x, v_t(y_i)) v_t(y_i).\label{eq:MCNO_estimator}
\end{equation}
This avoids spectral transforms or hierarchical architectures and achieves linear complexity in the number of sampled points. 
To enhance efficiency, MCNO leverages PyTorch’s \texttt{einsum} to perform the 
per-sample feature mixing in parallel by evaluating the kernel as
\[
\kappa_\phi(y_i, v_t(y_i)) = \phi_i v_t(y_i),
\]
where each $\phi_i \in \mathbb{R}^{d_v \times d_v}$ is a learnable tensor acting on the 
feature vector $v_t(y_i) \in \mathbb{R}^{d_v}$, and we write $\phi = \{\phi_i\}_{i=1}^N$ for the full set of kernel parameters.
This design enables efficient GPU 
parallelization and avoids costly per-sample matrix operations before the Monte 
Carlo aggregation step. The Monte Carlo–type Neural Operator (MCNO) achieves 
linear computational complexity with respect to the number of samples $N$, since 
kernel interactions are computed only at the sampled points. 

Because the Monte Carlo integral is evaluated on a subset of points, the resulting 
latent representation is interpolated to the full grid using lightweight linear 
interpolation, preserving efficiency while producing a structured output compatible 
with standard neural architectures. Although applied across the full grid, our 
reconstruction step scales linearly with $N_{\text{grid}}$ and involves only 
inexpensive local averaging, introducing only modest overhead compared to the 
kernel evaluation.

\begin{figure}[h!]
    \centering
    \includegraphics[width=0.8\textwidth]{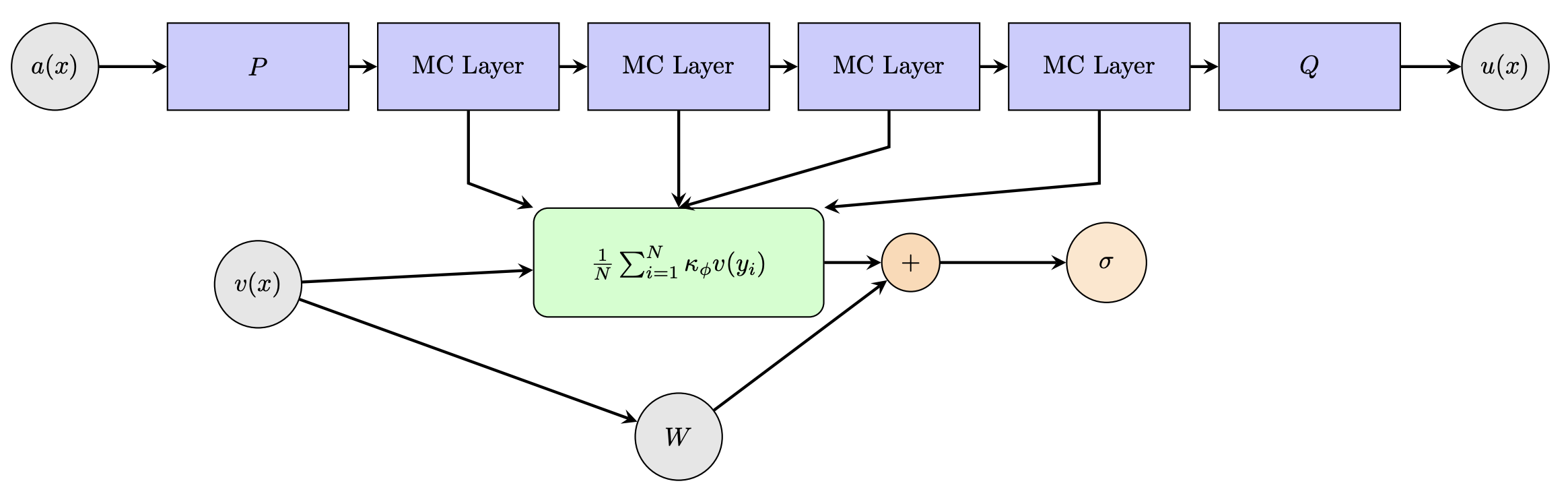}
    \caption{MCNO architecture.}
    \label{fig:Picture3}
\end{figure}
\begin{algorithm}[h]
\caption{MCNO — Forward Pass}
\begin{algorithmic}[1]
\STATE \textbf{Given:} parameters $\phi$, $W$; samples $\{y_i\}_{i=1}^N$; grid; activation $\sigma$; interpolation rule
\STATE \textbf{Input:} current representation $v_t$
\STATE \textbf{Output:} $v_{t+1}(x)$
\STATE Estimate kernel: $(\hat{\mathcal{K}}_{N,\phi}v_t)(x)=\tfrac{1}{N}\sum_{i=1}^N \kappa_\phi\!\big(x, v_t(y_i)\big)\, v_t(y_i)$
\STATE Interpolate $(\hat{\mathcal{K}}_{N,\phi}v_t)$ onto the full grid
\STATE Update: $v_{t+1}(x)=\sigma\!\left(W v_t(x)+(\hat{\mathcal{K}}_{N,\phi}v_t)(x)\right)$
\end{algorithmic}
\end{algorithm}

\section{Bias–Variance Analysis and Dimensional Scaling} 
At each layer, MCNO computes the kernel aggregation \eqref{eq:MCNO_estimator} using
$N$ points $\{y_{s_i}\}_{i=1}^N$ sampled uniformly from the grid
$\{y_j\}_{j=1}^{N_{\text{grid}}}\subset D\subset\mathbb{R}^d$.
Each sample–feature update is used once, so the per-layer cost is
$\mathrm{Cost}_{\text{MCNO}}=\mathcal{O}(N\,d_v)$ (feature width $d_v$).
In practice, tensor contractions parallelized over $x$ give aggregation
$\mathcal{O}(N)$, while interpolation/reconstruction scales with $N_{\text{grid}}$.

\paragraph{Error decomposition.}
Define
\begin{equation}
E(x)= (\hat{\mathcal{K}}_{N,\phi} v_t)(x)-(\mathcal{K}_\phi v_t)(x).
\end{equation}
Adding and subtracting
$(\mathcal{K}_\phi^{\text{grid}} v_t)(x)=\tfrac{1}{N_{\text{grid}}}\sum_{j=1}^{N_{\text{grid}}}\kappa_\phi(x,y_j)v_t(y_j)$ yields
\begin{equation}
E(x)=\underbrace{(\mathcal{K}_\phi^{\text{grid}} v_t)(x)-(\mathcal{K}_\phi v_t)(x)}_{\text{Bias}}
+\underbrace{(\hat{\mathcal{K}}_{N,\phi} v_t)(x)-(\mathcal{K}_\phi^{\text{grid}} v_t)(x)}_{\text{Variance}}.
\end{equation}

\paragraph{Bias.}
Partition $D$ into hypercubes $\{C_j\}_{j=1}^{N_{\text{grid}}}$ with side $h$ so $N_{\text{grid}}=(1/h)^d$.
For each $C_j$ centered at $y_j$,
$\int_{C_j}\kappa_\phi(x,y)v_t(y)\,dy=\kappa_\phi(x,y_j)v_t(y_j)h^d+R_j(x)$ with
$|R_j(x)|\le L\,\tfrac{\sqrt d}{2}\,h^{d+1}$, hence
\begin{align}
\big|(\mathcal{K}_\phi^{\text{grid}} v_t)(x)-(\mathcal{K}_\phi v_t)(x)\big|
&\le \sum_{j=1}^{N_{\text{grid}}}|R_j(x)|
\le L\,\tfrac{\sqrt d}{2}\,h\,N_{\text{grid}}h^d
= C_1 N_{\text{grid}}^{-1/d}. \label{eq:bias}
\end{align}

\paragraph{Variance.}
For fixed $x$, $(\hat{\mathcal{K}}_{N,\phi} v_t)(x)=\frac{1}{N}\sum_{i=1}^N Z_i(x)$ with
$Z_i(x)=\kappa_\phi(x,y_{s_i})v_t(y_{s_i})$, $|Z_i(x)|\le C$.
Hoeffding gives
\begin{equation}
\mathbb{P}\!\left(\big|(\hat{\mathcal{K}}_{N,\phi} v_t)(x)-\mathbb{E}[(\hat{\mathcal{K}}_{N,\phi} v_t)(x)]\big|>t\right)
\le 2\exp\!\left(-\frac{N t^2}{2C^2}\right).
\end{equation}
By a union bound over $\{x_j\}_{j=1}^{N_{\text{grid}}}$ and failure $\delta$,
\begin{equation}
\sup_{x\in D}\big|(\hat{\mathcal{K}}_{N,\phi} v_t)(x)-(\mathcal{K}_\phi^{\text{grid}} v_t)(x)\big|
\;\le\; C_2\sqrt{\frac{\log(2N_{\text{grid}}/\delta)}{N}}. \label{eq:variance}
\end{equation}

\paragraph{Sample complexity and cost.}
From \eqref{eq:bias}–\eqref{eq:variance},
\begin{equation}
\sup_{x\in D}|E(x)|\;\lesssim\; C_1 N_{\text{grid}}^{-1/d}+C_2\sqrt{\frac{\log N_{\text{grid}}}{N}}.
\end{equation}
To reach tolerance $\varepsilon$, choose
$N_{\text{grid}}=\mathcal{O}(\varepsilon^{-d})$ and
$N=\mathcal{O}(\varepsilon^{-2}\log(\varepsilon^{-d}))$.
Since aggregation is $\mathcal{O}(N)$ and reconstruction is $\mathcal{O}(N_{\text{grid}})$, the total cost is
\begin{equation}
\mathrm{Cost}_{\text{MCNO}}(\varepsilon,d)
=\mathcal{O}(N)+\mathcal{O}(N_{\text{grid}})
=\tilde{\mathcal{O}}(\varepsilon^{-2}+\varepsilon^{-d}),
\end{equation}
with $\tilde{\mathcal{O}}$ hiding logarithms.

\paragraph{High-dimensional note.}
Dimension $d$ enters via the bias (through $N_{\text{grid}}^{-1/d}$) and grid cost; the Monte Carlo
variance, and thus the stochastic sample complexity, remains dimension-independent.

\color{black}
\section{Numerical experiments}
We evaluate the Monte Carlo-type Neural Operator on two standard benchmark problems: Burgers' equation and Korteweg-de Vries (KdV) equation. These benchmarks are widely used for testing operator learning methods due to their diversity in  complexity and relevance to applications.
To ensure fairness and consistency, we adopt the same datasets and experimental setups as described in \cite{li2021fourier}.
The MCNO employs a four-layer architecture of Monte Carlo kernel integral operators, with ReLU activations. Training uses 1000 samples, with 100 additional samples reserved for testing. Optimization is performed with the Adam optimizer, starting at a learning rate of 0.001, halved every 100 epochs over a total of 500 epochs. We set the width to $64, \ N=100$ for Burgers' equation and $N=75$ for Korteweg-de Vries (KdV) equation. All experiments are conducted on an Tesla V100-SXM2-32GB of memory GPU and the accuracy is reported in terms of relative $L_2$ loss.
The  errors for the main benchmarks FNO, MWT and WNO are reported with our implementation on the same GPU that we used to train our model. For the other benchmarks: GNO, LNO and MGNO, the errors are taken from \cite{gupta2021multiwavelet}. These models were trained and tested on an Nvidia V100-32GB GPU, therefore the comparison with their reported errors is fair to a certain extent.

\subsection{Burgers' Equation: Dataset and Results}

We consider the one-dimensional Burgers' equation, a nonlinear PDE commonly used to model viscous fluid flow, with periodic boundary conditions and initial condition \(u(x,0)=u_0(x)\), where \(u_0 \in L^2_{\text{per}}((0,1);\mathbb{R})\) and \(\nu>0\) is the viscosity: 
\(\partial_t u + \partial_x (u^2/2) = \nu \partial_{xx} u\), \(x\in(0,1)\), \(t\in(0,1]\). Following the setup of the Fourier Neural Operator (FNO) benchmark \cite{li2021fourier}, the initial condition is sampled from a Gaussian \(\mu = \mathcal{N}(0,625(-\Delta+25I)^{-2})\), and the PDE is solved with a split-step method on a high-resolution grid (\(8192\) points) and subsampled for training and testing. The task is to learn the operator mapping \(G^\dagger: u_0 \mapsto u(\cdot,1)\). Table~\ref{tab:burgers} summarizes results of different neural operators across resolutions: the proposed Monte Carlo-type Neural Operator (MCNO) achieves competitive performance, outperforming GNO, LNO, MGNO, and FNO in speed and accuracy. While WNO performs best on higher resolutions, it is inconsistent across grids and slower than MCNO and FNO. The MWT Leg model achieves slightly lower consistent errors but is computationally expensive. Overall, MCNO offers a balanced trade-off between accuracy, efficiency, and adaptability, with relative $L_2$ error and per-epoch time remaining stable as the number of samples increases (Figures~\ref{fig:Picture6} and~\ref{fig:Picture7}).
\begin{table}[ht]
\centering
\caption{Benchmarks on 1-D Burgers’ Equation showing relative $L_2$ errors for different input resolutions $s$.}
\label{tab:burgers}
\resizebox{0.95\linewidth}{!}{%
\begin{tabular}{|l|c|c|c|c|c|c|c|}
\hline
\textbf{Model} & \makecell{Time per epoch \\ (s=256, s=8192)} & $s=256$ & $s=512$ & $s=1024$ & $s=2048$ & $s=4096$ & $s=8192$ \\ \hline
GNO~\cite{li2020neuralb}&- & 0.0555  & 0.0594  & 0.0651   & 0.0663   & 0.0666   & 0.0699   \\ \hline
LNO~\cite{lu2021learning}&- & 0.0212  & 0.0221  & 0.0217   & 0.0219   & 0.0200   & 0.0189   \\ \hline
MGNO~\cite{li2020multipolea}&- & 0.0243  & 0.0355  & 0.0374   & 0.0360   & 0.0364   & 0.0364   \\
\hline
WNO~\cite{tripura2022wavelet}&  (2.45s, 3.1s) & 0.0546   & 0.0213  &  0.0077  & 0.0043   &  0.0027   &  0.0012  \\ \hline
FNO~\cite{li2021fourier}& (0.44s, 1.56s) &  0.0183 &  0.0182 &  0.0180  &  0.0177  &  0.0172 & 0.0168   \\ \hline
MCNO      & (0.40s, 1.32s)  & 0.0064   &0.0067   & 0.0062   & 0.0069   & 0.0071    & 0.0065   \\ \hline
MWT Leg~\cite{gupta2021multiwavelet}& (3.70s, 7.10s) & 0.0027  & 0.0026  &  0.0023  &  0.0024  &  0.0025  & 0.0023   \\ \hline
\end{tabular}
}
\end{table}
\subsection{Korteweg-de Vries (KdV) equation: Dataset and Results}
\begin{figure}[h!]
    \centering
    \begin{minipage}[b]{0.48\textwidth}
        \centering
        \includegraphics[width=\textwidth]{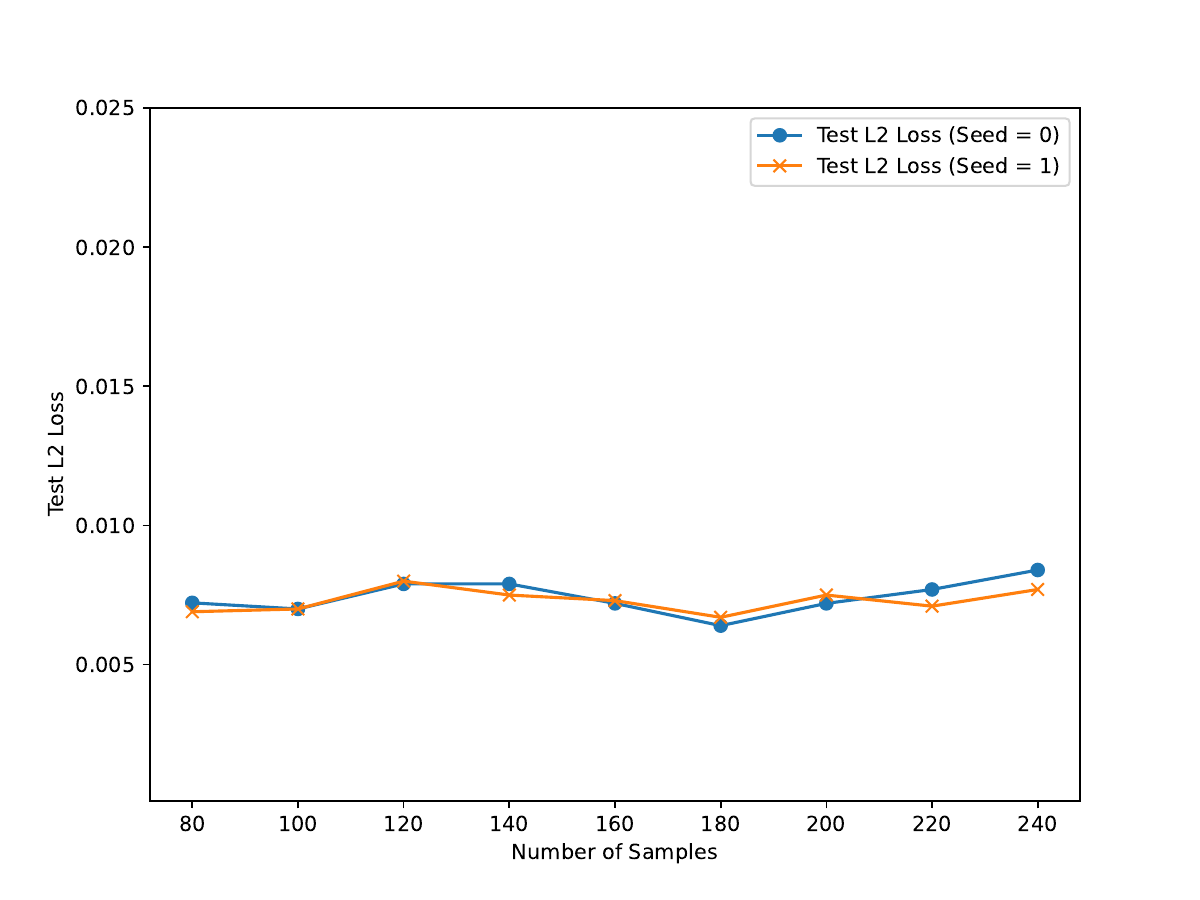}
        \caption{Relative $L_2$ test loss vs Number of samples for Burgers}
        \label{fig:Picture6}
    \end{minipage}
    \hfill
    \begin{minipage}[b]{0.48\textwidth}
        \centering
        \includegraphics[width=\textwidth]{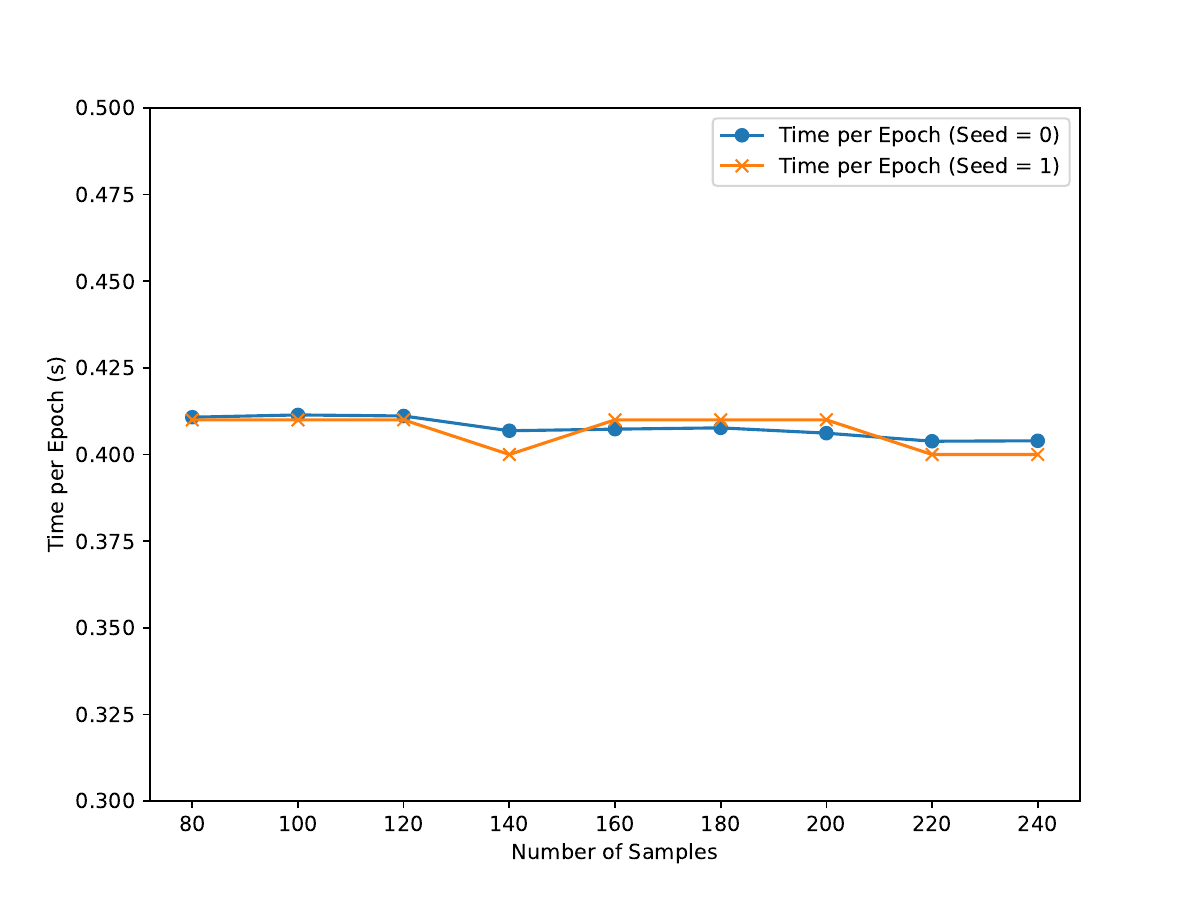}
        \caption{Time per epoch vs Number of samples for Burgers}
        \label{fig:Picture7}
    \end{minipage}

    \vspace{0.3cm} 

    \begin{minipage}[b]{0.48\textwidth}
        \centering
        \includegraphics[width=\textwidth]{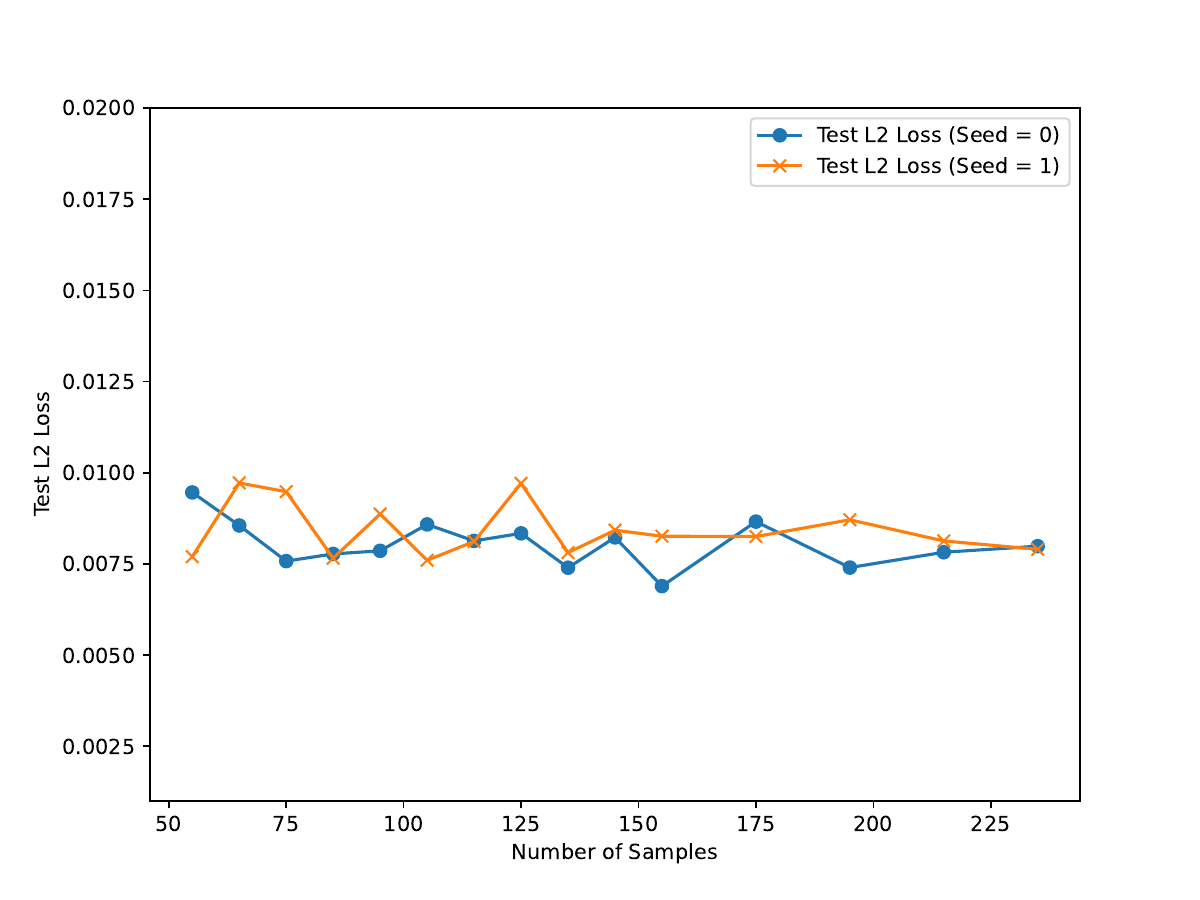}
        \caption{Relative $L_2$ test loss vs Number of samples for KdV}
        \label{fig:Picture4}
    \end{minipage}
    \hfill
    \begin{minipage}[b]{0.48\textwidth}
        \centering
        \includegraphics[width=\textwidth]{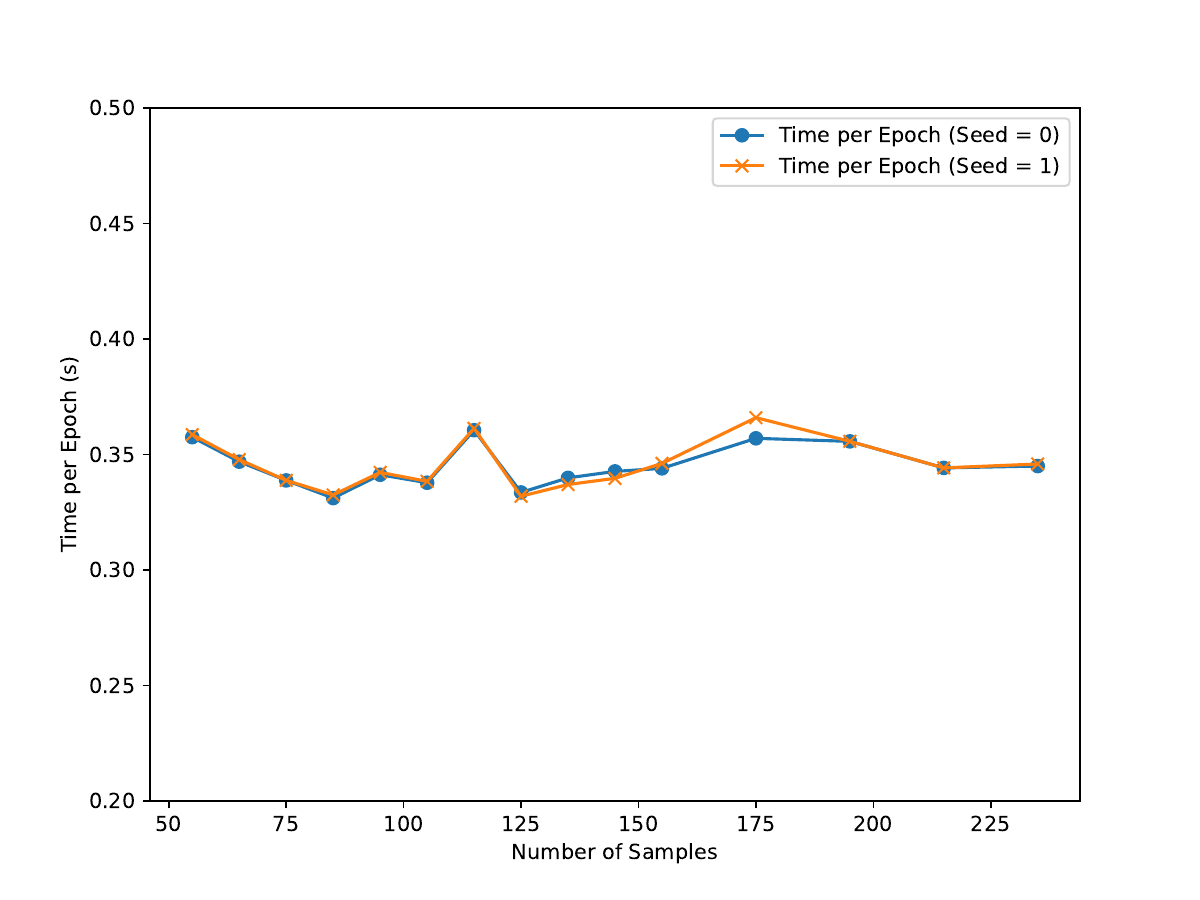}
        \caption{Time per epoch vs Number of samples for KdV}
        \label{fig:Picture5}
    \end{minipage}
\end{figure}

We consider the one-dimensional Korteweg-de Vries (KdV) equation, a nonlinear PDE modeling shallow water waves and dispersive phenomena, defined as \(\partial_t u = -0.5\, u \partial_x u - \partial_x^3 u\), \(x \in (0,1)\), \(t \in (0,1]\). Following the dataset setup in \cite{gupta2021multiwavelet}, the initial conditions \(u_0(x) = u(x,0)\) are sampled from Gaussian random fields with periodic boundary conditions, \(u_0 \sim \mathcal{N}(0,7^4(-\Delta+7^2 I)^{-2.5})\), and high-resolution solutions are computed with Chebfun (\(2^{10}\) points) and subsampled for training and testing. The task is to learn the operator mapping \(u_0 \mapsto u(\cdot,1)\). Table~\ref{tab:kdv_results} reports benchmark results across resolutions: the proposed Monte Carlo-type Neural Operator (MCNO) outperforms FNO, MGNO, LNO, and GNO, capturing the nonlinear dynamics effectively. While MWT Leg achieves slightly lower errors, MCNO balances accuracy and efficiency, with relative $L_2$ loss and per-epoch computation remaining stable as the number of samples increases (Figures~\ref{fig:Picture4} and~\ref{fig:Picture5}).



\begin{table*}[ht]
\centering
\caption{Korteweg-de Vries (KdV) equation benchmarks showing relative $L_2$ errors for different input resolutions $s$.} 
\label{tab:kdv_results}
\begin{tabular}{|l|c|c|c|c|c|}
\hline
\textbf{Model}    & \makecell{Time per epoch \\ (s=128, s=1024)}   & \textbf{$s=128$} & \textbf{$s=256$} & \textbf{$s=512$} & \textbf{$s=1024$} \\ \hline
MWT Leg      & (3.24s, 4.96s)    & 0.0036 & 0.0042  & 0.0042  & 0.0040     \\ \hline
MCNO &     (0.36s, 0.49s)    & 0.0070  & 0.0079   & 0.0088   & 0.0081  \\ \hline
FNO          &  (0.48s, 0.50s)   & 0.0126  & 0.0125   & 0.0122   & 0.0133     \\ \hline
MGNO         &  -   & 0.1515  & 0.1355   & 0.1345   & 0.1363       \\ \hline
LNO          &  -  & 0.0557  & 0.0414   & 0.0425   & 0.0447       \\ \hline
GNO          &  -  & 0.0760  & 0.0695   & 0.0699   & 0.0721       \\ \hline
\end{tabular}
\end{table*}

\section{Conclusion}

We introduced the Monte Carlo-type Neural Operator (MCNO), a lightweight framework that learns kernel functions using a single Monte Carlo sample to approximate integral operators. Unlike spectral or hierarchical approaches, MCNO does not assume translation invariance or rely on global bases, offering flexibility across grid resolutions and problem settings.
Experiments on one-dimensional PDEs, including Burgers’ and KdV equations, demonstrate that MCNO achieves competitive accuracy with computational efficiency and architectural simplicity. These results highlight MCNO as a practical alternative to established neural operator models.
Future work includes extending MCNO to higher-dimensional PDEs, exploring adaptive sampling strategies, and applying it to unstructured or heterogeneous domains.

\section*{Acknowledgment}

This work was supported by the NYUAD Center for Interacting Urban Networks (CITIES), funded by Tamkeen under the NYUAD Research Institute Award CG001. The views expressed in this article are those of the authors and do not reflect the opinions of CITIES or their funding agencies.
\bibliographystyle{plain}

\begin{thebibliography}{1}

\bibitem{gupta2021multiwavelet}
Hitesh Gupta, Kaushik Bhattacharya, and Anima Anandkumar.
\newblock Multiwavelet-based operator learning for differential equations.
\newblock {\em Proceedings of Machine Learning Research (PMLR)}, 139:12144--12155, 2021.

\bibitem{kovachki2023neural}
Nikola Kovachki, Zongyi Li, Burigede Liu, Kamyar Azizzadenesheli, Kaushik Bhattacharya, Andrew Stuart, and Anima Anandkumar.
\newblock Neural operator: Learning maps between function spaces with applications to partial differential equations.
\newblock {\em Journal of Machine Learning Research}, 24(89):1--97, 2023.

\bibitem{li2020multipolea}
Zongyi Li, Nikola Kovachki, Kamyar Azizzadenesheli, Burigede Liu, Kaushik Bhattacharya, Andrew Stuart, and Anima Anandkumar.
\newblock Multipole graph neural operator for parametric partial differential equations.
\newblock {\em arXiv preprint arXiv:2009.01938}, 2020.

\bibitem{li2020neuralb}
Zongyi Li, Nikola Kovachki, Kamyar Azizzadenesheli, Burigede Liu, Kaushik Bhattacharya, Andrew Stuart, and Anima Anandkumar.
\newblock Neural operator: Graph kernel network for partial differential equations.
\newblock In {\em Proceedings of the International Conference on Learning Representations (ICLR)}, 2020.

\bibitem{li2021fourier}
Zongyi Li, Nikola Kovachki, Kamyar Azizzadenesheli, Burigede Liu, Kaushik Bhattacharya, Andrew Stuart, and Anima Anandkumar.
\newblock Fourier neural operator for parametric partial differential equations.
\newblock {\em International Conference on Learning Representations (ICLR)}, 2021.

\bibitem{lu2021learning}
Lu~Lu, Pengzhan Jin, Guofei Pang, Zhongqiang Zhang, and George~Em Karniadakis.
\newblock Learning nonlinear operators via deeponet based on the universal approximation theorem of operators.
\newblock {\em Nature Machine Intelligence}, 3(3):218--229, 2021.

\bibitem{tripura2022wavelet}
Tapas Tripura and Souvik Chakraborty.
\newblock Wavelet neural operator for solving parametric partial differential equations in computational mechanics problems.
\newblock {\em Computer Methods in Applied Mechanics and Engineering}, 404:115783, 2023.

\end{thebibliography}

\end{document}